\documentclass{INTERSPEECH2023}
\usepackage{multicol}
\usepackage{multirow}

\usepackage{times}
\usepackage{epsfig}
\usepackage{graphicx}
\usepackage{amsmath}
\usepackage{amssymb}
\usepackage{booktabs}
\usepackage{bbm}
\usepackage{stfloats}
\usepackage{multicol}
\usepackage{multirow}

\usepackage{colortbl}
\usepackage{xcolor}
\usepackage{array, caption, threeparttable}
\usepackage{amsbsy}
\usepackage{tipa}
\usepackage{booktabs}

\usepackage[capitalize]{cleveref}
\usepackage{expl3}
\ExplSyntaxOn
\newcommand\latinabbrev[1]{
	\peek_meaning:NTF . {% Same as \@ifnextchar
		#1\@}%
	{ \peek_catcode:NTF a {% Check whether next char has same catcode as \'a, i.e., is a letter
			#1.\@ }%
		{#1.\@}}} 
\ExplSyntaxOff

\def\eg{\latinabbrev{e.g}}

\newcommand\blfootnote[1]{%
  \begingroup
  \renewcommand\thefootnote{}\footnote{#1}%
  \addtocounter{footnote}{-1}%
  \endgroup
}
\interspeechcameraready

\title{The Hidden Dance of Phonemes and Visage: Unveiling the Enigmatic Link between Phonemes and Facial Features}

\name{Liao Qu $^{1,*}$, Xianwei Zou $^{1,*}$, Xiang Li $^{1,*}$, Yandong Wen $^2$, Rita Singh $^1$, Bhiksha Raj $^{1,3}$}
\address{
  $^1$Carnegie Mellon University \\
  $^2$Max Planck Institute \\
  $^3$Mohamed bin Zayed University of Artificial Intelligence}
\email{\{liaoq,xianweiz,xl6,yandongw,rsingh,bhiksha\}@andrew.cmu.edu}

\begin{document}

\maketitle
 
\begin{abstract}

\end{abstract}
This work unveils the enigmatic link between phonemes and facial features. Traditional studies on voice-face correlations typically involve using a long period of voice input, including generating face images from voices and reconstructing 3D face meshes from voices. 
However, in situations like voice-based crimes, the available voice evidence may be \textit{short} and \textit{limited}. Additionally, from a physiological perspective, each segment of speech - \textit{phoneme} corresponds to different types of airflow and movements in the face.
Therefore, it is advantageous to discover the hidden link between phonemes and face attributes.
In this paper, we propose an analysis pipeline to help us explore the voice-face relationship in a fine-grained manner, i.e., phonemes \textit{vs.} facial anthropometric measurements (AM). We build an estimator for each phoneme-AM pair and evaluate the correlation through hypothesis testing. Our results indicate that AMs are more predictable from vowels compared to consonants, particularly with plosives. Additionally, we observe that if a specific AM exhibits more movement during phoneme pronunciation, it is more predictable. Our findings support those in physiology regarding correlation and lay the groundwork for future research on speech-face multimodal learning.

%%%%%%%%%%% 修改重写之后 %%%%%%%%%%%%
% Traditional studies on speech-face correlations typically involve using the entire speech, whether converting voices into face images or reconstructing 3D face meshes from voices. However, in situations like voice-based crimes, the available voice evidence may be \textit{short} and \textit{limited}. Therefore, it is advantageous to focus on the recorded signal's small segments- \textbf{\textit{phonemes}}, and discover the hidden link between phonemes and face features. 
%  Our work unveils the elusive relationship between phonemes and facial features. By focusing on phonemes, we can discover how each speech sound corresponds to different facial features, allowing for a more precise understanding of the speech-face relationship. In this paper, we propose a \textit{\textbf{phoneme-anthropometric measurements (AMs)}} paradigm, which enables us to explore the speech-face relationship in a fine-grained manner, i.e., phonemes \textit{vs.} distances/angles/proportions. We build an estimator for each phoneme-AM pair and evaluate the correlation through hypothesis testing. The results xxxxXxxxx, which can help us better understand how phonemes and facial features are interrelated. 
 
 % This shift towards exploring the relationship between phonemes and facial features represents an exciting advancement in the field, with potential benefits for various industries and applications.

\noindent\textbf{Index Terms}: voice-face correlation, phoneme

\blfootnote{$^*$ Equal contribution.}
\section{Introduction}

% 从自然科学的角度入手（两三句，引用），发现音频和脸存在关系，然后我们想研究更为细粒度的东西

% This work studies the correlation between phonemes and facial features 
The implicit relation between speech and anthropometry features has been extensively researched in recent years. 
% Numerous voice profiling studies \cite{Bahari2012AgeEF, McGilloway2000AutomaticRO, Ptacek1966AgeRF, Li2019ImprovingTS} have shown that human voice carries a plethora of information about the speaker and it is possible to infer the speaker's biophysical characteristics, such as gender, age, health conditions, etc. from their speech. 
Numerous voice profiling studies \cite{Bahari2012AgeEF, McGilloway2000AutomaticRO, Ptacek1966AgeRF, Li2019ImprovingTS, SnchezHevia2022AgeGC, Zhang2019AttentionaugmentedEM} have shown that human voice carries a plethora of information about the speaker, making it possible to deduce biophysical characteristics of speakers, \eg, gender, age and health conditions, from their voice.
% human voice and face contain highly correlated and mutual-complemented information
However, in criminal profiling scenarios, the study of correlations between voice and face becomes essential. In voice-based crimes, such as hoax emergency calls and voice-based phishing, the officers seek to depict the facial features of the criminal merely from \textit{short} voice evidence. ``Mayday" can be an example of the audio samples obtained by officers. This motivates us to investigate the phoneme-level correlation between voice and face.

Several recent works have attempted to investigate the correlation between voice and face. Cognitive science studies \cite{Belin2004ThinkingTV, Hardcastle1999TheHO} suggests human has a strong capability to imagine the appearance of speakers based on their voice. To verify it, face reconstruction from voice, which aims to recover identity-fidelity faces from their corresponding voice recordings, is introduced by \cite{Oh2019Speech2FaceLT}. After that, great progresses \cite{Choi2020FromIT, Wen2019FaceRF} has been achieved by using advanced Generative Adversarial Networks \cite{Goodfellow2014GenerativeAN}. Going beyond, recent works \cite{Wu2022CrossModalPC, Wu2021Voice2MeshC3} attempt to recover 3D face geometry meshes from voice to avoid the impact of inevitable background area modeling in 2D images. However, all these approaches rely on a long period of voice and potentially neglect the advantage of exploring a more fine-grained voice-face correspondence. 
% did not dig deep into which parts of the speech signal might be informative and deterministic to their prediction. 
 % Moreover, there might not be much voice evidence accessible in the case of many voice-based crimes. In these situations, it is especially advantageous to regard the speech signal as small  segments and make use of the most informative ones.

Rethinking the human voice production mechanism, the voice is produced by either the vibration of the vocal cord or the resonance of the pulmonary airflow. For both of the mechanisms, the vocal track is highly enrolled. The vocal track can be assumed as a filter, reflecting the characteristics of human voice. With the tight bind of muscle and bone, the vocal track is also correlated with facial attributes. Specifically, each phoneme corresponds to a different vocal track status and also an accordingly facial movement. To construct an accurate voice-face correlation, we argue that phoneme-level voice-face modeling is vital. 

To investigate and understand the voice-face correlation at a more fine-grained phoneme level, we propose an analysis pipeline that leverages a common feature extractor with a regression head to predict human anthropometric measurements (AM) from phoneme. Specifically, Human anthropometric measurements are a set of facial measurements summarized from cognitive science studies that can effectively represent the identity of a human. We decompose the audio recordings into phonemes and learn to predict AMs from phonemes. In this way, we can quantitatively analyze the relationship between each facial AM and phoneme pairs. In this paper, we aim to answer core two questions: 1) whether there exists any ``enigmatic" link between phoneme and facial features and 2) whether those ``enigmatic" links can be quantitatively described.

\begin{figure*}[!htb]
\begin{center}
% \fbox{\rule{0pt}{2in} \rule{.9\linewidth}{0pt}}
\includegraphics[width=1.0\linewidth]{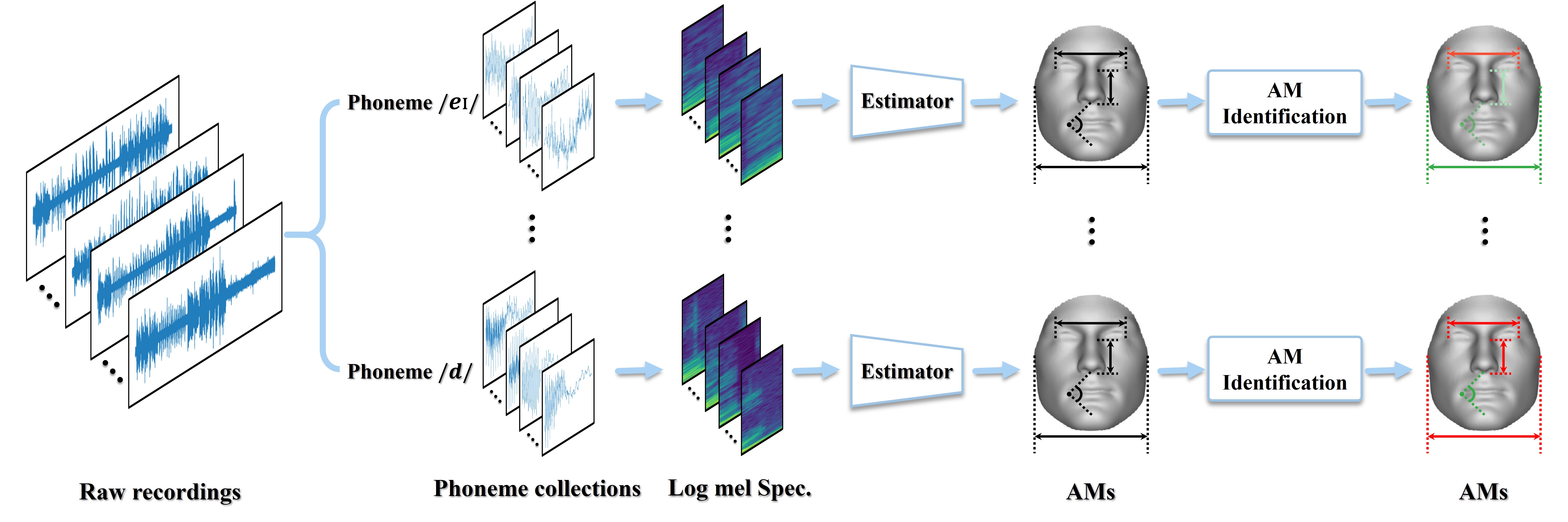}
\end{center}
% \vspace{-1.2em}
\caption{Illustration of our framework. We convert phoneme clips into mel spectrograms and develop estimators for each phoneme-acoustic model (AM) pair. Hypothesis testing is used to determine the predictability of AMs from phonemes. Green denotes predictable AMs and red otherwise, where the color shade indicates the degree of predictability.} % 我们用绿色表示了predictable的AM，红色表示不predictable，颜色深浅表示。。。
\label{fig:model}
% \vspace{-0.8em}
\end{figure*}
% \newpage

\section{Related Works}
\noindent\textbf{Learning Human Attributes from Voice.}
There is a substantial body of research on inferring human attributes from a person's voice, including speaker identity \cite{Bull1983TheVA, Ravanelli2018SpeakerRF}, age \cite{Bahari2012AgeEF, Ptacek1966AgeRF}, gender \cite{Li2019ImprovingTS}, and emotion status \cite{Wang2017LearningUR, Zhang2019AttentionaugmentedEM}. In addition to predicting attributes directly related to voice, many studies have explored the implicit correlation between voice and facial features. 
One popular task is generating 2D face images from voice using GANs \cite{Goodfellow2014GenerativeAN}, which has been progressed in several recent works \cite{Choi2020FromIT, Oh2019Speech2FaceLT, Wen2019FaceRF}. To avoid the impact of inevitable background area modeling in 2D images, recent work turns to the 3D domain: synthesizing 3D meshes from voices \cite{Wu2022CrossModalPC}.

\noindent\textbf{Phoneme Pronunciation Mechanism.}
The human vocal tract can be considered as a series of resonance chambers that can be dynamically configured \cite{riede2005vocal}. When the vocal cords vibrate, they convert the airflow from the lungs into acoustic energy in the form of sound waves, producing \textit{voice}. 
The shape and dimensions of the resonant chambers change as the movements of the vocal tract modify the acoustic signal, resulting in different patterns \cite{Singh2016ForensicAF, Ghiselin1974DarwinAF}. To produce a specific pattern, the mouth and nose must form a corresponding shape. Each pattern corresponds to a unique compositional unit of speech, or a \textit{phoneme}. When a speaker enunciates different phonemes, the vocal tract, mouse, nose, and other related facial structures act in concert. And each phoneme, therefore, carries some information about all these related features.

\section{Methods}
\subsection{Overview}
% 必须要写的3部分：切割phoneme，estimator，t检验
We aim to investigate the correlations between each phoneme and AM pair. As shown in Fig \ref{fig:model}, we first transform the segmented phonemes into log mel spectrum to better capture information from the frequency domain. After that, an AM estimator is employed to predict each AM from phonemes. Finally, we use hypothesis testing to analyze the correlation between each phoneme-AM pair.

% \vspace{1em}
% \subsection{Phoneme segmentation} % 标题斟酌一下
% To identify predictable AMs and their corresponding phonemes, the first step is to extract individual phonemes from the dataset. 
% % This requires us to analyze each recording and separate the phonemes. 
% However, due to the large amount of data and the complexity of distinguishing phoneme intervals, manually segmenting phonemes can be laborious, difficult, and imprecise.

% To improve the accuracy of phoneme segmentation, we employ state-of-the-art phoneme segmentation approaches. Specifically, we use the Wav2Vec2-Large-XLSR-53 model \cite{Xu2022SimpleAE} developed by FAIR, which learns powerful speech representations from more than 50.000 hours of unlabeled speech. This model is trained using a contrastive task on masked latent speech representations and can learn a quantization of the latent shared across languages. It is then fine-tuned on multi-lingual labeled common voice data. Since our data primarily contains standard English pronunciations, this model can provide relatively high segmentation accuracy.

% To extract the phonemes, we first load the \texttt{.wav} data and feed it into the Wav2Vec2CTC decoder. After decoding, we obtain the location of each phoneme and the gap between them. By mapping the decoded result back to the original data, we successfully generate the phoneme dataset.
\subsection{Notations}
% Now, our problem is defined as follows: we have a set of paired voice recordings of phonemes and AMs, and we want to predict each AM from different phonemes. After segmenting the recordings into phonemes, they can be represented as $\boldsymbol{P} = \{p^{(1)}, p^{(2)}, \dots, p^{(k)}\}$, where $k$ represents the total number of distinct phonemes. Similarly, the AMs can be represented as $AMs = \{m^{(1)}, m^{(2)}, \dots, m^{(n)}\}$, where $n$ is the number of summarized AMs. We refer to the entire dataset as $D$. To simplify the training and evaluation process, we divide $D$ into three subsets: a training set $D_t$ for estimator learning, a validation set $D_{v_1}$ for estimator selection, and a validation set $D_{v_2}$ for hypothesis testing and AM-phoneme pair selection.

Our problem involves a set of paired voice recordings of phonemes and AMs, where we aim to predict each AM from different phonemes. We begin by segmenting the recordings into phonemes, which can be represented as $\boldsymbol{P} = {p^{(1)}, p^{(2)}, \dots, p^{(k)}}$, where $k$ denotes the total number of distinct phonemes. Similarly, the AMs can be represented as $AMs = {m^{(1)}, m^{(2)}, \dots, m^{(n)}}$, where $n$ represents the number of summarized AMs. We refer to the entire dataset as $D$.

To simplify the training and evaluation process, we divide $D$ into three subsets. The first subset is the training set $D_t$, which is used for estimator learning. The second subset is the validation set $D_{v_1}$, which is used for estimator selection. Finally, the third subset is the validation set $D_{v_2}$, which is used for hypothesis testing and AM-phoneme pair selection.

\subsection{AM Estimator} % 标题斟酌一下

We leverage an AM estimator $E_{ij}$ to predict the $j$-th AM from the $i$-th phoneme $m^{(i)}=E_{ij}(p^{(j})$ as an estimator that maps the $j$-th phoneme to the $i$-th AM. To begin, we transform each phoneme into a log mel spectrum, which is essentially an image. This is a classic regression problem, and therefore, we need a model with strong feature extraction capabilities to extract information from the image.
We develop a modified version of the classical MNasNet model developed by Google AI. It is designed to be efficient, lightweight, and highly accurate for tasks such as image classification and object detection. \cite{Tan2018MnasNetPN}. Our modification retains the original structure, but with a few modifications to the input and output layers. Specifically, we change the input Conv2d module to accept only 1 channel, and the output Linear module to produce only 1 value. In addition, since this part is model-independent, other models such as ResNet \cite{He2015DeepRL} are also capable of achieving the same function.

\subsection{Hypothesis Testing for Phoneme-AM Predictability}
Once AMs are predicted from different phonemes, the next step is to determine whether a specific phoneme can actually predict an AM. To do this, we use hypothesis testing for each AM-phoneme pair separately.
 Firstly, we write the null hypothesis and the alternative hypothesis for the $i$-th AM and the $j$-th phoneme as 
\[H_{0}: \text{AM } m^{(i)} \text{ is not predictable from phoneme } p^{(j)}\]
\[H_{1}: \text{AM } m^{(i)}  \text{ is predictable from phoneme }p^{(j)}\]
\vspace{0.3em}

To reject the null hypothesis $H_{0}$, we need to compare our estimator $E_{ij}$ for the AM $m^{(i)}$ when using phoneme $p^{(j)}$ as input with a chance-level estimator $C_{ij}$. If the performance of $E_{ij}$ is statistically significantly better than $C_{ij}$, we can reject $H_{0}$ and accept $H_{1}$. To estimate the chance level for phoneme $p^{(j)}$ in our training set $D_{t}$, we use the mean $m^{(i)}$ of all instances of that phoneme in the set. Specifically, we calculate a constant value $C_{ij}$ as follows: $C_{ij} = \frac{1}{|D_{t}|} \sum_{m^{(i)} \in D_{t}} m^{(i)}$. We can express the hypotheses as:

\vspace{0.3em}
\[ H_0: \mu(\varepsilon_{ij}/\varepsilon_{ij}^C) \geqslant 1 \]
\[ H_1: \mu(\varepsilon_{ij}/\varepsilon_{ij}^C) < 1 \]
\vspace{0.3em}

Here, $\mu(\cdot)$ represents the mean function, and $\varepsilon_{ij}$ and $\varepsilon_{ij}^C$ are the mean squared errors (MSE) of the estimators $E_{ij}$ and $C_{ij}$ on the validation set $D_{v_2}$, respectively. We can compute them as follows:
\[ \varepsilon_{ij} = \frac{1}{|D_{v_2}|} \sum_{m^{(i)}\in D_{v_2}}(\hat{m}^{(i)}-m^{(i)})^2 \]
\[ \varepsilon_{ij}^C = \frac{1}{|D_{v_2}|} \sum_{m^{(i)}\in D_{v_2}}(C_{ij}-m^{(i)})^2 \]
\vspace{0.3em}

To conduct repeated experiments, we need to train the estimators multiple times. In each iteration, we randomly split the dataset into $D_t$, $D_{v_1}$, and $D_{v_2}$. We then use the one-sided paired-sample t-test to test the hypothesis. The confidence interval (CI) bounds are:
\[ CI_l = \mu \left( \frac{\varepsilon_{ij}}{\varepsilon_{ij}^C} \right) - t_{1-\alpha,\nu} \cdot \frac{\sigma(\varepsilon_{ij}/\varepsilon_{ij}^C)}{\sqrt{N}} \]

\[ CI_u = \mu \left( \frac{\varepsilon_{ij}}{\varepsilon_{ij}^C} \right) + t_{1-\alpha,\nu} \cdot \frac{\sigma(\varepsilon_{ij}/\varepsilon_{ij}^C)}{\sqrt{N}} \]
\vspace{0.3em}

Here, $\sigma(\cdot)$ represents the standard deviation function, $N$ represents the number of experiments, $\alpha$ represents the significance level, and $\nu = N-1$ represents the degree of freedom. For this project, we set $N=10$, and we choose $\alpha=0.05$ to obtain statistically significant results. We can read the value of $t_{1-\alpha,\nu}$ directly from the t-distribution table.
To test the hypothesis, if the CI upper bound $CI_u < 1$, we can infer that we successfully reject $H_{0}$ and accept $H_{1}$, meaning that the AM $m^{(i)}$ is predictable from phoneme $p^{(j)}$. On the contrary, if $CI_u \geq 1$, we cannot reject $H_{0}$, indicating that the result is not statistically significant.

\section{Experiments}
\subsection{Dataset}

We conducted experiments on a private audio-visual dataset $D$. The dataset contains 1,026 individuals' paired voice recordings and scanned 3D facial shapes. Each recording is a raw speech speaking out general phonemes and sentences with a length of 1-2 minutes. Each facial data consists of 6790 3D-coordinate points collected from one person. % 如何split数据集

\subsection{Data Processing and Training}

% 我们选取了>5000条数据量的phoneme
\begin{figure}[!htb]
\begin{center}
\includegraphics[width=1.0\linewidth]{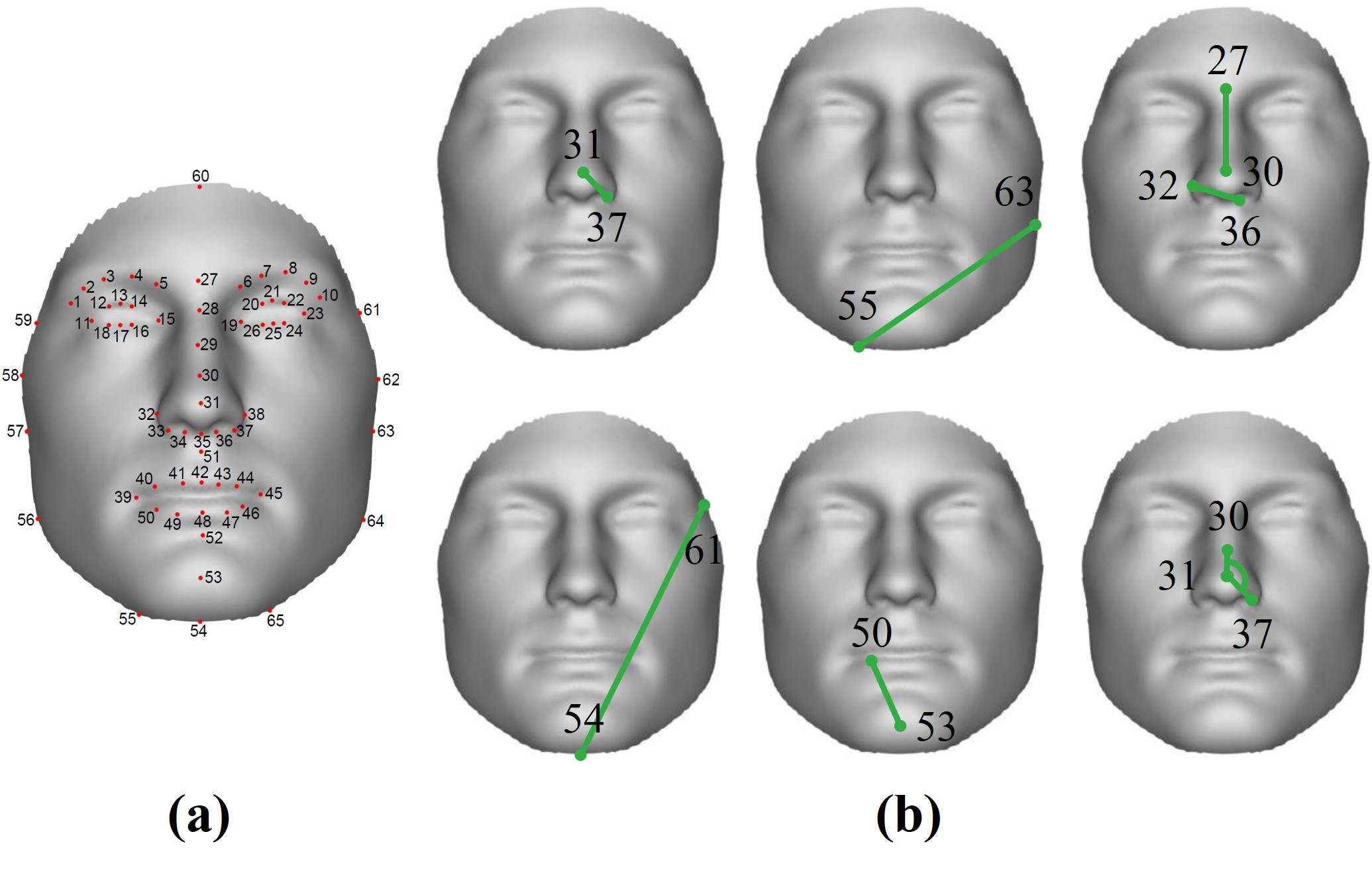}
\end{center}
% \vspace{-1.2em}
\caption{(a) The selected landmarks. (b) The visualization of the 6 most predictable AMs. They are arranged in descending order from left to right and top to bottom. Numbers in the face denote the index of landmarks.}
\label{fig:am-visualization}
% \vspace{-1.5em}
\end{figure}

\noindent\textbf{Phoneme segmentation.} % 标题斟酌一下
To identify predictable AMs and their corresponding phonemes, the first step is to extract individual phonemes from the dataset. 
% This requires us to analyze each recording and separate the phonemes. 
However, due to the large amount of data and the complexity of distinguishing phoneme intervals, manually segmenting phonemes can be laborious, difficult, and imprecise.

To improve the accuracy of phoneme segmentation, we employ state-of-the-art phoneme segmentation approaches. Specifically, we use the Wav2Vec2-Large-XLSR-53 model \cite{Xu2022SimpleAE} developed by FAIR, which learns powerful speech representations from more than 50.000 hours of unlabeled speech. This model is trained using a contrastive task on masked latent speech representations and can learn a quantization of the latent shared across languages. It is then fine-tuned on multi-lingual labeled common voice data. Since our data primarily contains standard English pronunciations, this model can provide relatively high segmentation accuracy.

We adopt the \texttt{wav2vec2-xlsr-53-espeak-cv-ft} in  huggingface \footnote{\textit{https://huggingface.co/facebook/wav2vec2-xlsr-53-espeak-cv-ft}} in our experiments. After splitting, we choose the most frequently used phonemes which have number of samples $\geq$ 5000. The detailed list is provided in the label of Fig. \ref{fig:corr}. For each phoneme recording, we follow \cite{Wen2019FaceRF} and perform 64-dimensional log mel-spectrograms using an analysis window of 25ms, with a hop of 10ms between frames. We perform normalization by mean and variance of each mel-frequency bin.

\noindent\textbf{AM summarization.}
We summarize the most commonly used AMs \cite{Ghafourzadeh2019PartBased3F, Shan2021AnthropometricA, Zhuang2010FacialAD, wen2022reconstruction, Farkas2004AnthropometricMO}, including distances, proportions, and angles in Table \ref{tab:summarize-am}. The selected landmark is shown in Fig. \ref{fig:am-visualization} (a). These AMs are more robust than 3D coordinate representations. This is attributed to the complete elimination of variations induced by spatial misalignment, thus rendering them more reliable and resistant to perturbations. The ground truth AMs are normalized to have a mean of zero and a variance of one.

\begin{table}[!htb]
  \centering
  \fontsize{9}{10}\selectfont
    \caption{The summarized AMs.}
    % \vspace{-0.5em}
    \label{tab:summarize-am}
    \begin{tabular}{ccc}
        \hline\toprule
        \multicolumn{3}{c}{\textbf{distance}} \\
        \midrule
        31-37 & 32-36 & 40-42 \\
        39-43 & 33-35 & 50-53 \\
        2-7 & 30-53 & 59-53 \\
        55-63 & 54-61 & \\
        \midrule
        \multicolumn{3}{c}{\textbf{proportion}} \\
        \midrule
        31-37 / 27-30 & 32-36 / 27-30 & 31-37 / 59-53 \\
        32-36 / 59-53 & 54-64 / 31-37 & 56-62 / 31-37\\
        \midrule
        \multicolumn{3}{c}{\textbf{angle}} \\
        \midrule
        31-30-37 & 31-29-37 & 29-30-34 \\
        \bottomrule
        \hline
    \end{tabular}
    % \vspace{-0.5em}
\end{table}

\noindent\textbf{Training details.}
For each phoneme-AM pair, we conduct 10 repeated experiments to ensure statistical significance. In each experiment, we randomly sample 5000 data samples and randomly split them into the $D_t/D_{v_1}/D_{v_2}$ set in the ratio of $70\%/10\%/20\%$. We follow the typical settings of Adam \cite{Kingma2014AdamAM} for optimization of the estimator. The loss function we use is the mean squared error loss. The size of the mini-batch and learning rate is set to 128 and 0.0001, respectively.
% All of our models are trained on a single NVIDIA RTX 3090 ti GPU.  

\subsection{Results}

\subsubsection{Analysis of phonemes}

\begin{figure}[!htb]
\begin{center}
% \vspace{-1cm}
\includegraphics[width=1.0\linewidth]{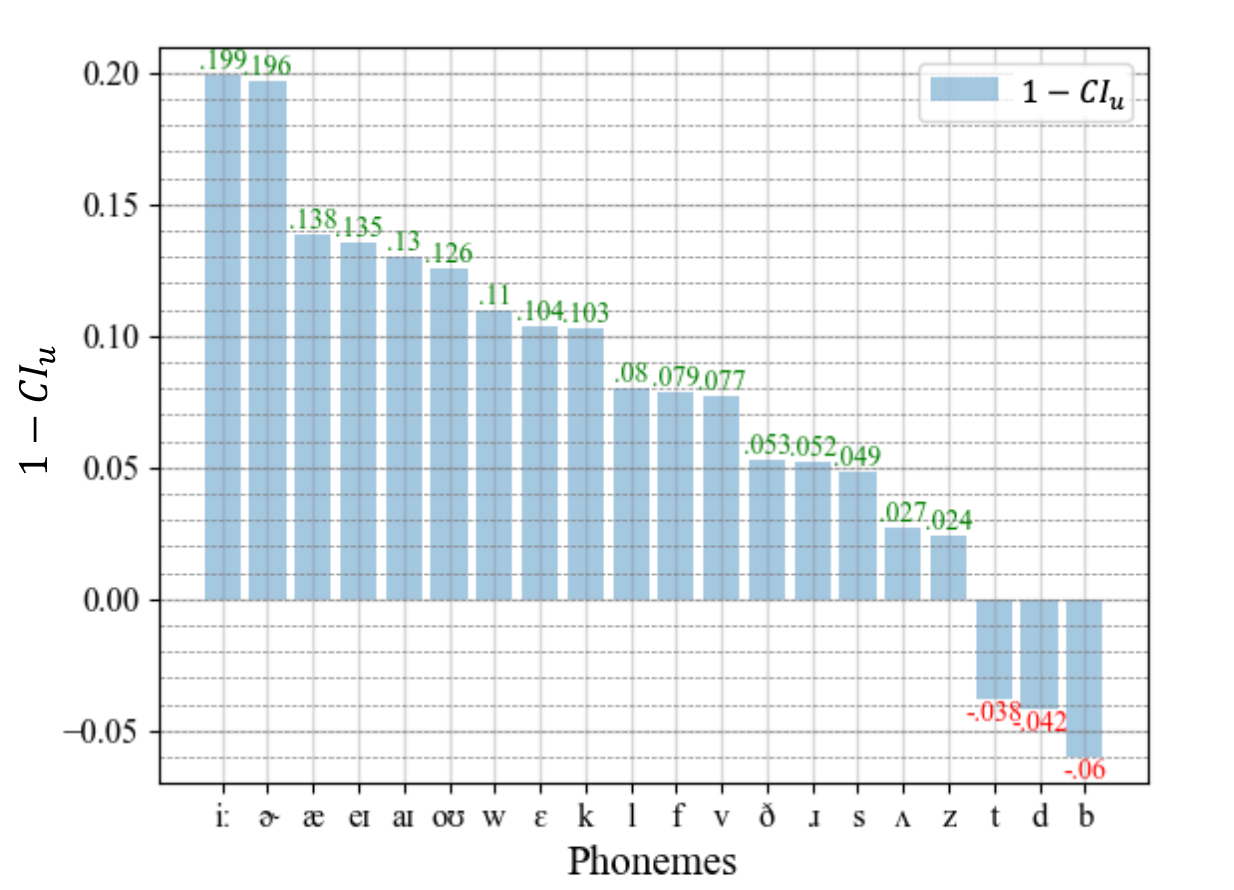}
\end{center}
\vspace{-1.2em}
\caption{Phonemes in descending order by avg. $1-CI_u$.}
\label{fig:corr}
\vspace{-1.5em}
\end{figure}

For each phoneme, we calculate the average $1 - CI_u$ result with every AMs. As can be seen from Fig. \ref{fig:corr}, \texttt{/\textipa{i:}/} got the highest avg. $1 - CI_u$ value 0.199, and \texttt{/\textipa{b}/} got the lowest value -0.06. 
% 说一下1-Clu小于0表示不可预测。as indicated in sec...
When $1 - CI_u$ is lower than 0, AMs are averagely unpredictable from the phoneme. 
The three phonemes with the lowest and negative values are \texttt{/\textipa{t}/}, \texttt{/\textipa{b}/} and \texttt{/\textipa{d}/}, which are all plosive consonants. During the pronunciation of plosive consonants, we complete stoppage of airflow followed by a sudden release of air through trivial mouse open and close, and there is minimal movement of the facial muscles and structures. Consequently, the prediction of any acoustic model based solely on such phonemes is challenging. On the contrary, most vowels achieve good performance in the test set, and all the top 6 phonemes belong to vowels with $1 - CI_u > 0.10$. Compared with consonants, there is no constriction of airflow in the vocal tract when pronouncing vowels. In order to produce specific vowels, the facial muscles have relatively greater movement during the pronunciation of these phonemes, such as jaw movement due to mouth opening or lip spreading. Thus vowel phonemes may carry more information about facial features. This, therefore, can make the model better capture the hidden correlation when predicting AMs.

\subsubsection{Analysis of AMs}
Similarly, for each AM, we also calculated the average $1-CI_u$ results with all phonemes. To intuitively locate the most predictable AMs (with the highest avg. $1-CI_u$) on the 3D face, we visualize them in Fig. \ref{fig:am-visualization} (b). Most of the predictable AMs are around the nose and mouth. On the contrary, AMs around the eyes are less predictable. This is consistent with the fact that the nose and mouth shapes (distances, proportions, and angles) affect the pronunciation of phonemes. Other than the nose and mouth, the jaw is another region frequently occurring in the most predictable AMs. 
Since the jaw is another region that exhibits frequent movement during pronunciation, we hypothesize that for a specific AM, if it is more frequently moved during the pronunciation of phonemes, the AM is generally more predictable. We further verify this hypothesis in the next section.

\subsubsection{Relationship between phonemes and AMs}

\begin{table}[!htb]
  \centering
  \fontsize{9}{10}\selectfont
    \caption{Detailed results of phoneme-AM pairs.}
    % \vspace{-0.5em}
    \label{tab:phoneme-am-pair}
    
    \fontsize{10}{12}\selectfont
    \renewcommand{\arraystretch}{1}
    \renewcommand{\tabcolsep}{1mm}
    \begin{tabular}{cccccccc}
        \hline\toprule
        AMs & \texttt{/\textipa{E}/} & \texttt{/\textipa{D}/} & \texttt{/\textipa{f}/} & \texttt{/\textipa{i:}/} & \texttt{/\textipa{v}/} & \texttt{/\textipa{w}/} & \texttt{/\textipa{\ae}/} \\
        \midrule
        39-43 & 0.10 & -0.04 & 0.08 & \textcolor{green}{0.23} & 0.11 & \textcolor{green}{0.18} & \textcolor{green}{0.18} \\ % (22)
        31-30-37 & 0.10 & 0.04 & \textcolor{green}{0.19} & 0.11 & 0.10 & \textcolor{green}{0.21} & -0.09 \\ % (89)
        50-53 & 0.05 & 0.09 & -0.07 & \textcolor{green}{0.21} & 0.06 & 0.11 & \textcolor{green}{0.21} \\ % (42)
        2-7 & 0.02 & 0.08 & 0.05 & 0.04 & -0.03 & 0.08 & 0.09 \\ % (7)
        \bottomrule
        \hline
    \end{tabular}
    % \vspace{0.5em}
\end{table}

\label{sec:relationship-between-phoneme-am}
We investigate the detailed relationship between phoneme and AM pairs to verify our hypothesis. As shown in Table \ref{tab:phoneme-am-pair}, we list 4 typical AMs paired with 7 phonemes, where 39-43 is an oblique distance of the lip, 31-30-37 is an angle in the nose, 50-53 is the distance between the lip and jaw, and 2-7 is the distance between eyebrow. In the case of AM 2-7, no matter pairing with any phoneme, the value is relatively low (all $1-CI_u$ values close to $0$). During the pronunciation process of any phonemes, the movement of this particular region is very limited. Therefore, phonemes can barely carry information about AMs in this part. However, for 39-43, it shows that \texttt{/\textipa{i:}/}, \texttt{/\textipa{w}/}, and \texttt{/\textipa{\ae}/} have the highest value. When pronouncing these three phonemes, the mouth usually grins in order to control the output airflow. And the distance between facial landmarks 39 and 43 could slightly influence the airflow from a physical perspective, therefore the phoneme produced may have subtle differences. In contrast, when pronouncing \texttt{/\textipa{f}/} and \texttt{/\textipa{D}/}, this AM barely moves. Thus, hardly can this AM influences the output airflow. The results verify that it is less predictable for these phoneme-AM pairs. Similarly, the phenomenon also occurs in other AMs like 50-53. It is more predictable when pairing with phonemes than needing mouse opening. All these experiments verify that for a specific AM, if it is more frequently moved during the pronunciation of phonemes, then the AM would be more predictable.

% \vspace{0.5em}
\section{Conclusions}
% 未来展望：使用更为精确的segmentor，使用predictable的完成人脸重建

In this work, we delve deeply into a fundamental question: whether there exists any ``enigmatic" link between phoneme and facial features. If so, whether those ``enigmatic" links can be quantitatively described? As a forerunner in this field, we design a phoneme-AMs paradigm, which enables us to explore the speech-face relationship in a fine-grained manner. Hypothesis testing is utilized to verify whether an AM is predictable from a phoneme. Experiments show that AMs are averagely more predictable with vowels compared with consonants, especially plosives, and are consistent with the physiological explanation. On the other hand, most of the predictable AMs are around the nose, mouth, and jaw. Results also verify that for a specific AM, if it moves more frequently during phoneme pronunciation, the AM will be more predictable since the phonemes might carry this hidden information during pronunciation. We hope our work lays a foundation for this field. In the future, we would like to scale up the range of phonemes and AMs to discover more hidden relationships. Moreover, we are investigating models that make use of the found hidden correlation knowledge in scenarios like 3D face reconstruction.

% This work unveils the enigmatic link between phonemes and facial features. Traditional studies on speech-face correlations typically involve using the entire speech, including converting voices into face images or reconstructing 3D face meshes from voices. 
% However, in situations like voice-based crimes, the available voice evidence may be \textit{short} and \textit{limited}. Therefore, it is advantageous to focus on the recorded signal's small segments- \textbf{\textit{phonemes}}, and discover the hidden link between phonemes and face features.  Our work unveils the elusive relationship between phonemes and facial features. By focusing on phonemes, we can discover how each speech sound corresponds to different facial features, allowing for a more precise understanding of the speech-face relationship. In this paper, we propose a \textit{\textbf{phoneme-anthropometric measurements (AMs)}} paradigm, which enables us to explore the speech-face relationship in a fine-grained manner, i.e., phonemes \textit{vs.} distances/angles/proportions. We build an estimator for each phoneme-AM pair and evaluate the correlation through hypothesis testing. The results xxxxXxxxx, which can help us better understand how phonemes and facial features are interrelated.

% \section{Acknowledgements}

% % \ifinterspeechfinal
% %      The INTERSPEECH 2023 organisers
% % \else
% %      The authors
% % \fi

% \newpage

\bibliographystyle{IEEEtran}
\bibliography{mybib}

\end{document}